# Classifying Objects in 3D Point Clouds Using Recurrent Neural Network: A GRU LSTM Hybrid Approach

Ramin Mousa[a], Mitra Khezli[b], Mohamadreza Azadi, Vahid Nikoofard,, Saba Hesaraki[e]

[a]Raminmousa@znu.ac.ir

[b]fm.khezli@gmail.com

[c]mohamadazadi33@gmial.com

[d] vahid@fat.uerj.br

[e]saba.hesaraki@srbiau.ac.ir

**Abstract**

Accurate classification of objects in 3D point clouds is a significant problem in several applications, such as autonomous navigation and augmented/virtual reality scenarios, which has become a research hot spot. In this paper, we presented a deep learning strategy for 3D object classification in augmented reality. The proposed approach is a combination of the GRU and LSTM. LSTM networks learn longer dependencies well, but due to the number of gates, it takes longer to train; on the other hand, GRU networks have a weaker performance than LSTM, but their training speed is much higher than GRU, which is The speed is due to its fewer gates. The proposed approach used the combination of speed and accuracy of these two networks. The proposed approach achieved an accuracy of 0.99 in the 4,499,0641 points dataset, which includes eight classes (unlabeled, man-made terrain, natural terrain, high vegetation, low vegetation, buildings, hardscape, scanning artifacts, cars). Meanwhile, the traditional machine learning approaches could achieve a maximum accuracy of 0.9489 in the best case.

## 1. Introduction

In recent years, Covid-19 has been changed our lifestyle. Virtual Reality (VR), Augmented Reality (AR), and mixed Reality (MR) has become a part of our lives [1].

We will propose the combined system, based on voice command, creating a VR environment using point cloud. The idea comes from the J.A.R.V.I.S. (Just A Rather Very Intelligent System) refers to fictitious personality, which is created by Marvel Cinematic Universe [3]. Jarvis helps Iron man to conduct his thoughts based on voice command. Our systems will create the environment by point cloud [6] and then recognize the object using a camera and add it into VR environment. Most interaction in a virtual environment depends on using hands to do critical tasks, which is more practical. Also, a hands-free interface based on voice recognition is followed by the eye and head gaze [4]. The role of Non-verbal and body language has been ignored in video communication. Although the remote assistant role is progressively increased, "due to a lack of spacial information, a limited field of view, a lack of context, and a limited transmission of non-verbal cues and body language, the affordances provided by video communication are insufficient [5]." Our combined system in this regard will concentrate on the role of voice and non-verbal communication in virtual reality. It will be a combination of verbal and non-verbal at the same time. In most articles [4, 5], only one area is covered, but it seems that combining them will give the user more freedom of choice. The role of the point cloud is deniable immersive virtual reality experiences in terms of simplicity and versatility is famous for presenting "photorealistic volumetric reconstructions of dynamic real-world objects." Also, user-centred Adaptive approach is a sensible way to increase the quality of bitrates [6].

## 2. Literature Review

In this section, we will assess the recent papers in virtual reality, including voice command, non-verbal command, point cloud techniques to create an immersive virtual environment. In the end, we will discuss some datasets tools that we would use for our proposed ideas.

### 2.1. Hands Free Interaction

Monteiro et al. [4] investigated the literature review usability of hands-free interaction techniques for VR environment for interaction tasks. They claimed that these techniques are valuable for users' interaction in grasping and directly interacting with objects. Most of the studies they examined used command voice and voice recognition techniques to do their tasks. The second popular method refers to eye tracking, integrated into head-mounted displays (HMD) and delivers accurate eye data for selection purposes. Similarly, head tracking methods are practical, especially for collecting head rotation data. They highlighted that verbal interaction is significantly increased for system control and selection tasks. In terms of evaluation metrics (Satisfaction, Efficiency, and Efficacy), most studies collected data from customized questionnaires of user feedback. Interaction time is an important measurement to evaluate the performance, while accuracy and number of errors are usually used for efficacy. Other metrics such as sense of presence and simulator sickness have been influenced the VR experience [4].

Binti Azizo et al. [9] proposed the "Virtual Reality 360 Universiti Teknologi Malaysia (UTM) Campus Tour with voice commands using the Head Mounted Display (HMD)", which is developed by using Unity3D and IBM Watson Speech Platform application programming interface (API). To decrease the expenses, they used smartphones. The primary function is voice command to control the virtual environment, which is a neutral way to communicate with a computer. They used three methods to evaluate their work: "black-box testing, white-box testing, and user usability." Also, there is some limitation on their work, including limitation due to the Covid-19 pandemic, weakness in internet connection, that will not allow voice recognition system has a good performance, being noisy environment will not allow the voice recognition functionally to detect users' voice, repetition in testing to get a good result. They suggested using the paid version of voice recognition software to overcome these limitations.

### 2.2. Point Clouds

The point clouds' advantages outweigh 3D meshes due to no lack of features, resilience to noise, no expensive preprocessing. However, it needs compression to store and transport efficiently over limited bandwidth networks. Subramanyam et al. [6] proposed the user-centred approach to generate the dataset based on navigation patterns in 6 degrees of freedom. Their evaluation system is based on user-centred, adaptive streaming to be objective quality and bitrate savings the delivery of dynamic cloud sequences through independently decodable tiles for real-time applications. To address optimization issues, adaptive utilize streaming over HTTP could be beneficial using the tiling strategy. Their experience was conducted using 26 participants, which each participant was requested to view 10 seconds dynamic point cloud and navigate within the scene freely. In terms of user navigation patterns, it indicated significant variation across sequences based on user movements on the XY floor. They observed "a greater spread of viewport locations around the object for this sequence [6] ."

The proposed rendering systems are based on a multi-pass rendering pipeline that allows users to explore the virtual environments by keeping visual quality and frame rates. They used 6DOF to avoid motion sickness. They used kd-tree for preprocessing techniques and their interaction handler is categorized into configuration and selection methods to apply rendering techniques, movement, distance measurements area, altering generated 3D point clouds. Their locomotion techniques refer to freely moving by the user, including: "Real Walking," "joystick flying," "point teleport (PT)," "dashing," or "locomotion based on gamepads and keyboards [7]."

### 2.2.1. *Generative Adversarial Network for the Point Cloud Generation*

In this section, we present a complete study of various generating models, including generative adversarial networks (GANs), that work on raw point clouds. GNA networks are very useful in this field because of their strong performance to create data similar to the distribution of input data. There are different models for generatingpoint clouds. Among these models, deep generative learning has been used the most. Deep generative models seek to integrate the interpretable representations provided by potential models into the scalability and flexibility of deep learning. In general, most machine learning models are discriminatory models [10]. Discriminatory models do not care how data is generated. They categorize the input data. In contrast, gener- ating models specify how data is generated to classify input data. In general, GANis a supervised learning framework that simultaneously training two sub-models: the G-Generative model, which attempts to create new instances of training data, and the D-discriminative model, which attempts to classify as real (from the training data field) or fake (created). G and D are both trained by playing in a zero-sum game. In particular, G tries to produce as many new samples as possible and maximizes the likelihood of misdiagnosis and D responsibility is to distinguish between real and fake examples. In the GAN network, the overall goal is to solve a two-person minimax problem. In the following, we will discuss some GAN studies in the field of raw point cloud production [11]. Three-dimensional geometric data provide excellent scope for studying productive representation and modelling learning. In an article [12], the authors deal with geometric data represented as point clouds. They have introduced an in-depth AutoEncoder(AE) network with advanced refurbishment quality and generalizability. Their proposed model consists of several main parts. In the first part, there is a GAN network that works directly with 3D point cloud data. This network is implemented within AE. Their AE network input is a 2048 dot cloud (a 2048 * 3 matrix) that represents a three-dimensional shape. Their encoder architecture is 1-dimensional convolutional layers with a core size of 1 and an increasing number of features. This approach encrypts each point independently. A "symmetric" variable change function (for example, a maximum reservoir) is placed after the twists to create a common representation. They used 5 1-dimensional convolutional layers, each with a ReLU and batch normalization. The output of the last convolution layer is transferred to a multi-layer fully connected to produce an output of 2048 * 3. Their proposed GAN was applied to the input of the set of raw points 2048 * 3, which was obtained through AE. The discriminator architecture is the same as AE (ie, from the kernel point of view, filter size and a number of neurons), the output of the last layer being completely connected to a sigmoid neuron. On the other hand, their GAN model generator takes a Gaussian noise vector as input and maps it to the output of 2048 * 3 through 5 layers of FC-ReLU. Their proposed approach on D-FAUST reached Performance 0.96.

Using tree-GAN structures can also provide acceptable results for 3D point cloud representation and generation. In [13] to achieve advanced performance for the multiclass 3D point cloud generation, a tree-structured graph convolutional network (Tree-GCN) is introduced as a generator for Tree-GAN. Because Tree-GCN uses the convolutional network in a tree, it can use ancestral information to enhance the display power of features. The Tree-GAN structure proposed by them consists of two networks, dis-criminator and generator. In the generator, a single point with a Gaussian distribution is considered as an input, and in each layer of this generator, GraphConv and Branching are located to produce a set of points. All points created in the previous layers are added to a tree in the current layer in order. The tree starts with the root node of z, divides it into child nodes through the Branching operation, and changes the nodes with the GraphConv operation. Finally, the generator generates a set of 3D-points as output. The Discriminator also shows the difference between the points produced and the real points. The network achieved FPD 0.809 and 0.439 on Chair and Airplane, respectively. Tree-GAN structures have been used in many other studies, the most important of which are models PT2PC [14], HSGAN [15], TreeGCN-ED [16], and SP-GAN [17] were mentioned.

In addition to their success in a point cloud, GANs have also been successful in Augmented Reality (AR). In AR, like Point Cloud, having methods to create data streams similar to model input streams can reduce expenses of data collection consumed-time. The following are some of the efforts made in the GAN composition for AR. Augmented Reality (AR) brings immersive interactive experiences in which the real and virtual worlds are highly interconnected. GAN networks have played a key role in AR. For example, studies can be provided: Providing a GAN-based model for linking different images of flammable and hazardous objects to assist firefighters [18], A model for mobile augmented reality (AR) [19], background augmentation generative adversarial networks (BAGANs) [20] with the aim of solving the problem of insufficient and asymmetric training data in AR object detection can be mentioned.

### *2.3. Object Detection*

Du et al. [21] proposed the 3D object detection framework named AGO-Net using point cloud and LiDAR camera from the actual scene by applying conceptual mode via domain adaptation. The distinction between object classification and localization based on abstract features categorization is sufficient for extracting details from point clouds. The network will be able to make connections between components for objects adaptively if the gap between perception and conception features is lost.

## 3. Methodology

### *3.1. Proposed Solution*

The basic idea underlying the proposed approach is explained below. We map the extracted features to a distributed vector (hence the encoding phase) and it is used to classify point cloud classes. In this section, after a brief overview of the notation and conventions we used, the various components of the network we tested will be described in detail. The proposed model is shown in Figure 1.

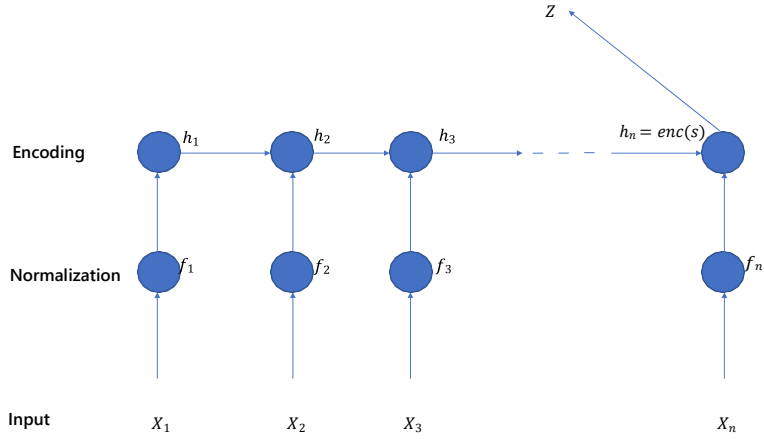

Figure 1: The proposed model

We used uppercase letters like $W$ and $U$ to represent matrices, lowercase letters like $b$ and $x$ to represent vectors and $x = [x^{<1>}, x^{<2>}, \ldots, x^{<n>}]$ to represent a vector of the input component. The index $i$ represents the i-the component of this input component. We will indicate the set of parameters of our model with the capital $\Theta$.

**Input layer:** Each record of feature is displayed as a vector $x = [x^{<1>}, x^{<2>}, \ldots, x^{<n>}]$. Where the i-the index of this vector represents the i-the property of this input component. We normalized the input because of the varying numerical scale of the input components. **The encoding layer:** A single distributed vector must record the meaning of the input. By having an input component s containing the h attribute, the result of the encoding phase output can be expressed as $enc(s) = h^{<k>}$, where $h^{<k>}R^j$ and the value of $j$ is a hyperparameter. We used a recurrent layer to encode the input layer. In this layer, the amount activated in the hidden layer depends on the current input value and the output value in the previous step. In general, we will have:

$$h^{<k>} = g(r^{<k>}, h^{<k-1>}, \theta_{enc})$$ (1)

Where g is the recurrent cell, $r^{<k>}$ the current input feature, $h^{<k-1>}$ the output of the hidden layer at time $k$, and $h^{<k-1>}$ is the output of the hidden layer at time $k - 1$ and $_{enc}$ are the learnable parameters in the learning phase. Accordingly, the encoding phase is as follows:

$$enc(S) = h^{<k>} = g(r^{<k>}, h^{<k-1>}, \theta_{enc})$$ (2)

$enc(S)$ requires recurrent layers to produce. An overview of the proposed return layers is shown in Figure 2.

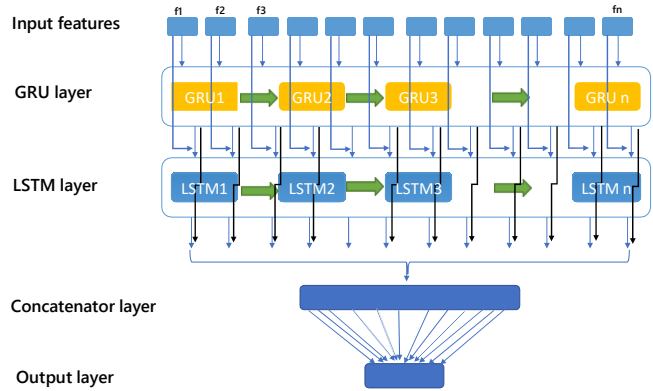

Figure 2: The proposed model

In fact, the proposed method is a four-layer structure. These four layers at a glance are:

1. **Input layer:** In this layer, each of the input features that are related to a classification are given to the GRU inputs.

2. **GRU layer:** The inputs of the GRU layer are vectors derived from the input layer.

3. **LSTM layer:** The inputs of the LSTM layer are vectors derived from the GRU layer.

4. **Output layer:** The output of the LSTM network is initially flattened.

Here are the details of each of these steps:

**GRU layer:** After the preprocessing operation is performed on the research data set, the data is sent to the GRU layer in the form of a normalized window. In this step, the number of GRU layer blocks was equal to the number of features. Figure 4 shows how this process works. In Figure 3, the first layer is the input layer and features and the second layer is the GRU layer. This figure shows the sending of samples to the GRU layer. The inputs of the GRU layer are vectors obtained through sliding windows and the output is calculated through the following equations.

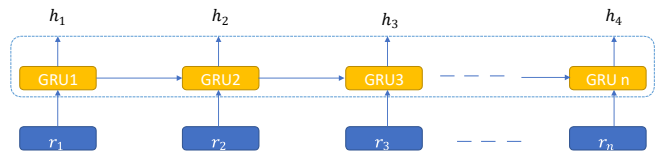

Figure 3: An overview of GRU

$$z_t = \sigma(w_s x_t + U_s h(t-1)) + b_z \qquad (3)$$

$$r_t = \sigma(w_r x_t + U_r h_t - 1) + b_r$$

$$h^{\mid} = \sigma(w_h x_t + U_n(r_t \odot h(t-1)) + b_h$$

$$h_t^t = (1-z)h_t - 1 + z_t h^{\mid}$$

Where $z_t$ is update gate, $r_t$ is rest gate, $h^{\mid}$ is candidate gate, and $h_t$ is output activation. $W_Z$, $W_R$, $W_h$, $U_Z$, $U_R$, $U_N$ are learnable matrixes, $b_n$, $b_s$, $b_r$ are learnable biases, $\sigma$ is sigmoid activation function, and $\odot$ is an element-wise multi-plication.

**LSTM layer:** The next step in the proposed method is to send the output of the GRU layer as input to the LSTM layer. Figure 4 shows how this process works.

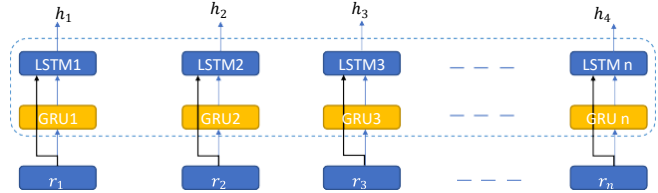

Figure 4: An overview of LSTM

The LSTM input in step $T$ is the vector $X \in R^E$. The hidden vector sequence is in LSTM, is calculated by the following equations:

$$f^{<k>} = \sigma(W_f\, r^{<k>} + U_f\, h^{<k-1>} + b_f) \tag{4}$$

$$q^{<k>} = \sigma(W_i\, r^{<k>} + U_i\, h^{<k-1>} + b_i)$$

$$g^{<k>} = tanh(W_g r^{<k>} + U_g h^{<k-1>} + b_g)$$

$$o^{<k>} = \sigma(W_o r^{<k>} + U_o h^{<k-1>} + b_0)$$

Where $i$ the input gate, $o$ the output gate, $f$ the forget gate, and $g$ the update gate, $[W_i, W_f, W_g, W_o, b_i, b_f, b_g, b_o]$ is the set of parameters to be learned. $q^{<k>}$ is updated through the following relationship:

$$q^{<k>} = f^{<k>} \odot q^{<k-1>} + i^{<k>} \odot g^{<k>} \tag{5}$$

where the $\odot$ symbol represents the element-wise product between two vectors. Finally, the activation of the cell is accomplished through the following relationship:

$$h^{<k>} = o^{<k>} \odot tanh\,(q^{<k>}) \tag{6}$$

To determine the class of each point, it is sufficient to apply a sigmoid to the encoding phase output as follows:

$$z = sigmoid(W_s E + b_s)x \tag{7}$$

### 3.2. Point Cloud Data

Finding a data set that could handle all the details and conditions for the proposed models was a difficult task in terms of collection. Therefore, the Point Cloud Segmentation data set was used in this work. This data set contains 44990641 points. Also, the initial design of this data set has been collected in 7 classes such as: Unlabeled, man-made terrain, natural terrain, high vegetation, low vegetation, buildings, hard scape, scanning artefacts, and cars. This data set contains information: x, y, z, intensity, r, g, b, and class; and is available through the link[1].

---

[1] https://www.kaggle.com/code/kmader/point-cloud-overview/data

### 3.3. Used Tools and Software

The Keras[2] has been used for developing the proposed GRULSTM model. The Keras is a high-level library written in Python. This API has the ability to run in seamlessly on both GPUs and CPUs environments. Keras is compatible with Python 2.7-3.x and provides various modules such as neural layers, cost functions, optimizers, initialization schemes, activation functions, and regularization. This API can use any of TensorFlow[3], CNTK[4], and Theano[5] backends. Using Keras is easier than it's backends, but does not mean that its flexibility is diminished.

### 3.4. Preparing Point Cloud Data

In this step, the initial preprocessing are performed on the training and test data. These preprocessing are performed in the same way in all datasets (train and test), and include two main steps in the following order:

- **Normalization:** We normalized the input because of the varying numerical scale of the input components. We use Linear Scale Transformation (Max-Min)[22]. for this purpose. We calculate the normalized value for each index of the input component according to the following equation:

$$r_{ij} = \frac{x_{ij} - x_j^{min}}{x_j^{max} - x_j^{min}} \tag{8}$$

- **Sliding Window:** In this step, sliding windows are formed for forecasting. Figure 3 shows how this process works. For example, for predicting $X_1 1$ time, all previous values from $X_1$ to $X_{10}$ are considered as input, and similarly, for predicting $X_{12}$, previous values from $X_2$ to $X_{11}$ are considered as input. This operation is used for all data, including training and test data. The remarkable thing about this model is that time can be used for hours, days and weeks.

---

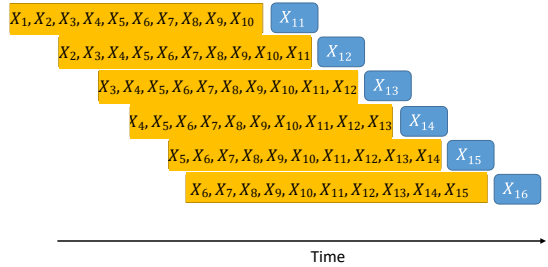

Figure 5: Sliding window process.

# 4. Result and discussion

## 4.1. Baseline models

1. **Gradient Boosting Classifier**: This method is used to develop classification and regression models to optimize the model learning process, which is primarily non-linear and is more commonly known as decision or regression trees. Regression and classification trees, individually, are poor models, but when considered as a set, their accuracy is greatly improved. Therefore, the sets are built gradually and incrementally so that each set corrects the error of the previous set mathematically in the form of the following equation[23]:

$$f_k(x) \quad = \sum_{m=1}^{k} \gamma_k \, h_m \, (x) \tag{9}$$

2. **Support Vector Machine(SVM)**:The purpose of this classifier is to make a hypothesis or train a model that predicts the class labels of unknown data or validation data samples that consist only of features. This model tries to find the largest margin between the data by creating a hyperplane. SVM kernels are generally used to map non-linear separable data into a higher-dimensional feature space sample consisting only of features[24]:

$$f(x) = sign(\Sigma_{j \in S \, V} \alpha_i *_i .k(x_i, x_{s \, V}) + b*)$$ (10)

Where K is the kernel, this kernel in our training data model is RBF, which is defined as follows:

$$K(x, y) = exp(-\frac{\|x - y\|^2}{2\sigma^2})$$ (11)

3. **XGBoost** : XGBoost [25] is widely used by data scientists to achieve advanced results in many machine learning challenges. The main idea of this algorithm is to present a new algorithm with dispersion awareness for scattered data, and a quantitative scheme for approximate tree learning. XGBoost has a very high predictive power, which makes it the best option for accuracy in various events because it can be used in both linear and tree models. This algorithm is approximately 10 times faster than existing gradient upgrade algorithms. This algorithm includes various objective functions, regression, classification and ranking. XG-Boost works as follows: If, for example, we have a $DS$ dataset that has $m$ attributes and $n$ instances of $DS = (h_i, y_i) : i = 1, ..., n, h_i \in \mathrm{R}^m, y \in \mathrm{R}.$ $_{pred}$ The $y^i$

predicted output is a group tree model produced by following equations[26]:

$$y^i_{pred} = \Sigma^k_{k=1} f_k (h_i)$$ (12)

Where $K$ represents the number of trees. $f_k$ represents (k-th tree). This model need to find the best set of functions by minimizing the loss and regularization objective according to following equation:

$$\ell(\phi) = \Sigma_i l(y^i, y^i_{pred}) + \sum_k \Omega(f_k)$$ (13)

Where $l$ the loss function and $\Omega$ is a measure of the complexity of the model and can help prevent the model from over-fitting. This criterion is obtained using the following equation:

$$\Omega(f_k) = \Upsilon T + \frac{1}{2}\lambda||W||^2 \qquad (14)$$

where $T$ represents the number of leaves and $w$ is the weight of each leaf in the decision tree. In decision trees, to minimize the objective function, function amplification is used in the model training process, which is used by adding a new function $f$ as a continuation of the model training. Therefore, in iteration $t$, a new function is added as follows:

$$\iota(t) = \sum_{i=1}^{n} l(y^i, y_{Pred}^i t - 1 + f_t (h_i))a^{n-k} + \Omega(f_t) \qquad (15)$$

$$\ell(split) = \frac{1}{2}\left[\frac{(\Sigma_{i \in I_L} g_i)^2}{\Sigma_{i \in I_R} + H_i + \lambda} + \frac{(\Sigma_{i \in I_R} g_i)^2}{\Sigma_{i \in I} + H_i + \lambda} - \frac{(\Sigma_{i \in I} g_i)^2}{\Sigma_{i \in I_L} + H_i + \lambda}\right] + \gamma \qquad (16)$$

$$g_i = \partial_{y_{pred\,t-1}^i} l(y_i, y_{pred\,t-1}^i) \qquad (17)$$

$$H_i = \partial_{\bar{y}_{t-1}^2}^2 l(y_i, y_{pred\,t-1}^i) \qquad (18)$$

4. **Random forests(RF)**: RF consist of a set of decision trees(DT)[27]. Mathematically, let $\hat{C}_b(x)$ be the class prediction of the bth tree; the class obtained from the random forest $\hat{C}_{rf}(x)$ is defined as follows:

$$\hat{C}_{rf}(x) = majorityvote\hat{C}_b(x)_l^B \qquad (19)$$

5. **Decision tree(DT)**: DT are powerful and popular tools used for both classification and prediction tasks. A DT represents rules that can be understood by humans and used in knowledge systems such as databases. These classification systems are in the form of tree structures. One of the most important questions that arise in decision tree-based models is how to choose the best split. The data set used is assumed to be a sample representation of real data, in which case reducing the error on the training data set can reduce the error on the test data. For this purpose, an attribute should be selected for the split that causes the separation of training samples of each class as much as possible, in other words, it causes the child nodes with less impurity. The following three different criteria can be used for this purpose[28][29]:

- Gini
- Entropy

Table 1: hyperparameters for different approaches.

| Model | Hyper parameters |
|---|---|
| Gradient Boosting Classifier | n estimators=3000, learning rate=0.05, max depth=4, subsample=1.0, criterion='friedman mse', min samples split=2, min samples leaf=1 |
| XGB Classifier | learning rate=0.1, n estimators=200, nthread=8, max depth=5, subsample=0.9, colsample bytree=0.9 |
| SVM | C=1.0, kernel='rbf', degree=3, gamma='scale', coef0=0.0, shrinking=True, tol=0.0001 |
| Random Forest | n estimators=100 |
| Decision Tree | n estimators=100 |
| LSTM | LSTM Block=100, Dropout=0.5, layers(LSTM(100),Dropout(0.5),Dense(100),Dense(8)) |
| GRU | GRU Block=100, layers(GRU(200),Dense(100),Dense(8)) |
| GRU+LSTM | GRU Block=100, LSTM Block=100, layers(GRU(100), LSTM(100), Dense(100) ,Dense(8)) |

| Model | Accuracy | Precision | Recall | F1-Score |
|---|---|---|---|---|
| Gradient Boosting Classifier | 0.8775 | 0.8627 | 0.9207 | 0.8904 |
| SVM | 0.8823 | 0.9212 | 0.9178 | 0.9194 |
| XGB Classifier | 0.9183 | 0.9234 | 0.9118 | 0.9175 |
| Random Forest | 0.9345 | 0.9347 | 0.9336 | 0.9341 |
| Decision Tree | 0.9489 | 0.9535 | 0.9524 | 0.9529 |

· Miss classification error

The following Table 4.1 are the hyperparameters of these models.

Table 2 shows the results of traditional machine learning approaches on the target data set. These approaches need to extract features manually. The Gradient Boosting Classifier approach in these data reached Accuracy=0.8775, Precision=0.8627, Recall=0.9207, and F1-Score=0.8904. The SVM model obtained better results than GB. This model was able to achieve Accuracy=0.8823. The superiority of this model in other metrics could also be considered, and it achieved Precision=0.92, Recall=0.9178, and F1=0.9194. XBG Classifier obtained higher accuracy than the two examined approaches, but in terms of F1 and Recall metrics, it obtained weaker results than SVM. This approach achieved Precision=0.9234, Recall=0.9118, and f1-score=0.9175. The Random Forest and Decision Tree approaches achieved the best results among them. These two approaches were able to reach a high accuracy of 0.93%. The Random Forest approach achieved Accuracy=0.9345, Precision=0.9347, Recall=0.9336, and F1=0.9341, respectively, and the Decision Tree approach also achieved Accuracy=0.9489, Precision=0.9535, Recall=0.9524, and F1-score=0.9529. acquired.

### 4.2. Proposed models

Table 2 shows the results of the proposed approach on the data set. These three approaches were compared in four criteria of evaluation of Accuracy, Precision, Recall and F1. All approaches have achieved a high accuracy of 0.99%. The GRU approach achieved an accuracy of 0.9989%, which is lower than the other two approaches. The LSTM approach also achieves an accuracy of 0.9990. Which was less hybrid than the proposed approach and better than the GRU. The difference between these approaches in the f1 criterion seems much better. The proposed approach has been able to achieve f1 = 0.9452%. All of these algorithms were implemented in 10 epochs and 10 blocks, 3000 batch size, and 10 fully connected layer size were used for each approach. In fact, it was tried to consider equal conditions for each of the algorithms.

Table 2: The results of applying the proposed models on point cloud dataset.

| Model | Accuracy | F1 | Recall | Precision |
|-------|----------|--------|--------|-----------|
| GRU | 0.9989 | 0.9294 | 0.9049 | 0.9182 |
| LSTM | 0.9990 | 0.9354 | 0.9495 | 0.9240 |
| GRULSTM | 0.9991 | 0.9420 | 0.9425 | 0.9452 |

Diagrams of different approaches to different performances are given in Figure 8. According to the accuracy chart, it can be concluded that all three models do not have overfitting. On the other hand, according to the error diagram and their logarithmic shape, it can be concluded that underfitting did not occur in all three models.

The remarkable thing about deep learning models is their parameter space. This space is continuous, which requires the correct selection of values. Two critical parameters in these networks are batch size and learning rate: Batch Size: Batch size determines the number of samples that a neural network uses in one session to train. The selection of categories has a balance between processing power and model learning speed. Learning rate: The learning rate shows the speed of convergence of an algorithm in deep learning. Incorrect selection of this value can trap the model in local minima.

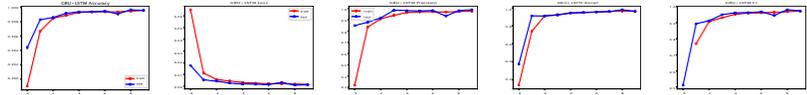

(a) GRULSTM Ac-   (b)    GRULSTM   (c)    GRULSTM   (d) GRULSTM Re-   (e) GRULSTM F1
curacy            Loss                   Precision         call

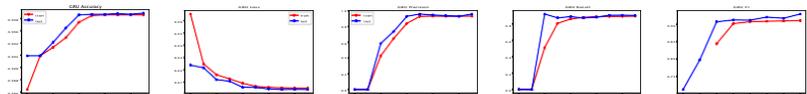

(f) GRU Accuracy   (g) GRU Loss   (h) GRUPrecision   (i) GRU Recall   (j) GRU F1

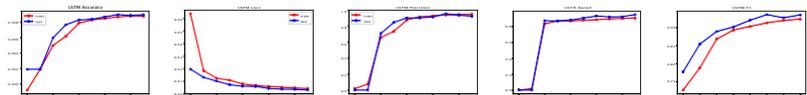

(k)   LSTM  Accu-   (l) LSTM Loss   (m)   LSTM  Preci-   (n) LSTM Recall   (o) LSTM F1
racy                                sion

Figure 6: Accuracy, loss, precision,recall, and f1 of different approach.

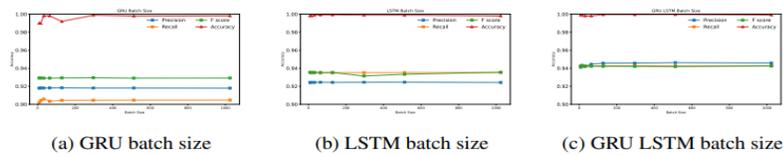

(a) GRU batch size      (b) LSTM batch size      (c) GRU LSTM batch size

Figure 7: Accuracy, loss, precision,recall, and f1 of different approach.

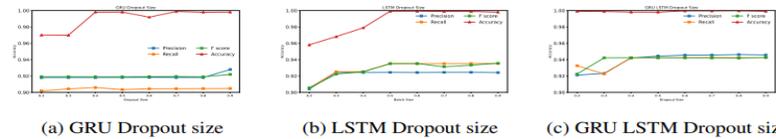

(a) GRU Dropout size    (b) LSTM Dropout size    (c) GRU LSTM Dropout size

Figure 8: Accuracy, loss, precision,recall, and f1 of different approach.

Figures 3 and 4 show the effect of choosing different categories and dropouts. The values [8,16,32,64,128,296,512,1024] were considered for the batch size, and [0.2,0.3,0.4,0.5,0.6,0.7,0.8,0.9] for the dropout size. The best results for models with high batch sizes and the worst results in Dropout are obtained for models with low rates.

## 5. Conclusion

In this article, point cloud classification in 3D space was investigated. Machine learning approaches have achieved poor results due to manual feature selection. These approaches achieved a maximum accuracy of 0.94 in large amounts of data and features with little data. On the other hand, deep learning approaches achieved an accuracy of 0.99 due to the automatic selection of features and the huge amount of training data. The challenge of speed and accuracy was raised in the proposed models. The LSTM approach has high accuracy and slow learning speed, and the GRU approach has a higher learning speed and lower learning accuracy than LSTM. For this purpose, it was tried to use the GRULSTM hybrid approach for classification. A fundamental issue in deep and machine learning approaches is that despite the high accuracy values, the models still have low F1-Score values. In the data set under study, some classes have a low frequency, and for this purpose, imbalance learning approaches can be used. The approaches [30][31][32][33] are suitable for this purpose, and these approaches can be used for future works. Also, the parameter space of the models is still challeng- ing, which can be overcome for future works by using PSO, Bat optimizer and APPE approaches to get the optimal parameter space.